\documentclass[a4paper]{article}

\usepackage{iwslt18,amssymb,amsmath,epsfig}

\usepackage{times}
\usepackage{latexsym}
\usepackage{url}

\usepackage[english]{babel}
\usepackage[utf8x]{inputenc}

\usepackage{hyperref} 
\usepackage{graphicx}
\usepackage{float}
\usepackage[justification=justified]{caption}
\usepackage{enumitem} 
\usepackage{dirtytalk}
\usepackage{array}
\usepackage{multirow}

\usepackage{siunitx}
\usepackage{color}

\newcommand{\citet}{\cite}
\newcommand{\citep}{\cite}

\title{Data Selection with Feature Decay Algorithms Using an Approximated Target Side}

 \makeatletter
 \def\name#1{\gdef\@name{#1\\}}
 \makeatother
 \name{{\em Alberto Poncelas, Gideon Maillette de Buy Wenniger, Andy Way}}

\address{ADAPT Centre, School of Computing, \\ Dublin City University, Dublin, Ireland\\
{\tt \{firstname.lastname\}@adaptcentre.ie}
}

\begin{document}

\maketitle

\begin{abstract}

Data selection techniques applied to neural machine translation (NMT) aim to increase the performance of a model by retrieving a subset of sentences for use as training data. 

One of the possible data selection techniques are transductive learning methods, which select the data based on the test set, i.e. the document to be translated. A limitation of these methods to date is that using the source-side test set does not by itself guarantee that sentences are selected with correct translations, or translations that are suitable given the test-set domain. Some corpora, such as subtitle corpora, may contain parallel sentences with inaccurate translations caused by localization or length restrictions.  

In order to try to fix this problem, in this paper we propose to use an approximated target-side in addition to the source-side when selecting suitable sentence-pairs for training a model. This approximated target-side is built by pre-translating the source-side. 

In this work, we explore the performance of this general idea for one specific data selection approach called Feature Decay Algorithms (FDA).

We train German-English NMT models on data selected by using the test set (source), the approximated target side, and a mixture of both.
Our findings reveal that models built using a combination of outputs of FDA (using the test set and an approximated target side) perform better than those solely using the test set. We obtain a statistically significant improvement of more than 1.5 BLEU points over a model trained with all data, and more than 0.5 BLEU points over a strong FDA baseline that uses source-side information only.

%We obtain a statistically significant improvements over a strong FDA baseline that only uses source-side information.

%We obtain a statistically significant improvement of more than 1 BLEU point over a strong FDA baseline that only uses source-side information.

\end{abstract}

\section{Introduction}
%One of the goal of the models in Machine Learning field, is to learn a model from a set of labeled points (training data) so it can accurately predict the label of a new, unlabeled, point. Having more data seems to be beneficial to build a model, however larger sets of data may entail problems such as the increase of noise or a more difficulty to generalize over all labeled points.

Supervised machine learning aims to learn predictive models from a set of labeled examples (training data) so that it can accurately predict the labels of new, unlabeled, examples. Having more data may seem at first glance to be beneficial to building more accurate models, but upon closer inspection this is not necessarily always the case. Machine learning models by design have an inductive bias that forces them to generalize over the training examples rather than just memorizing them without generalization. This means, however, that if the size of the training set is increased, this may lead to optimizing the model for predicting the labels of more examples, but which on average are less relevant at test time than would be the case for a more focused, smaller training set. The intuition of the importance of using a highly relevant set of training examples is captured well by the K-nearest neighbour model, which essentially computes at test time on-the-fly a very localized density estimate for every test example, based on the K training examples closest to the test example. It then uses this density estimate for classification. For the K-nearest neighbour model, increasing K too much is at the expense of basing predictions on an increasing number of less relevant examples. Furthermore similar to the K-nearest neighbour model, other predictive models which typically discard the original training examples and keep only a learned generalization over these examples, can suffer if the training data becomes bigger but on average less relevant to the test set.

%Other machine learning modeļ types, in contrast to K-nearest neighbour, are typically required to make predictions for all examples using a single model. But for these models having too many training examples that are further away from examples similar to the test set will consequently make the predictive model less specific to examples similar to the test set, analogous to having too big a K in K-nearest neighbour models. For this reason, larger sets of data may not increase performance, when the increased size coincides with decreases relevance to the test set.
%}

%however larger sets of data may entail problems such as the increase of noise or a more difficulty to generalize over all labeled points.

In Machine Translation (MT), the data used to build the models are parallel sentences (pairs of sentences in two languages, which are translations of each other) and we encounter the same problem when the amounts of data become excessively large. Too much training data may cause the model to be too generic, and not perform well if $test_{src}$ (the document to be translated, i.e. the test set), belongs to a specific domain.

Data selection techniques aim to solve that problem by selecting a subset of training data. Models that are trained on a small set of parallel sentences can perform better than those trained on all training data \citep{van2017dynamic,poncelas2018feature}. 

Within the data selection field we can find several approaches to reduce the data: select sentences of good translation quality (\textit{data quality}), select sentences relevant for a particular domain (\textit{domain adaptation}), or select sentences that are relevant for $test_{src}$ (\textit{transductive learning}). We focus on this last type, and so in this paper we propose new methods to build Neural Machine Translation (NMT) models that are tailored towards a $test_{src}$.

Transductive learning \citep{Vapnik1998} aims to find the best training instances given an unlabeled example. In MT this means finding the best parallel sentences given a document $test_{src}$ to be translated. In our work, the transductive data-selection method that we explore is Feature Decay Algorithms (FDA) \citep{biccici2011instance,bicici2015parfda,biccici2015optimizing}.  
%These techniques 
Standard FDA uses the {\em n}-grams of $test_{src}$ to retrieve training sentence pairs with source-side most similar to $test_{src}$. 
%The implicit assumption is that sentences with similar source-side also have similar target-side
FDA has demonstrated good performance in Statistical Machine Translation (SMT) and NMT \citep{poncelas2018feature}.

In most cases, FDA is used as a single step in the pipeline of building a model, using $test_{src}$ to extract a subset of parallel sentences. In this paper, we propose a different configuration of use of FDA for building NMT models (see left side of Figure \ref{src_trg_FDA_pipeline}). In particular, we propose executing FDA not only using the $test_{src}$ (source-side language), as is common, but additionally on a pre-translated test set (approximated target-side). In order to avoid confusion, in this work we use $test_{src}$ to indicate the test set (in the source-side language) and $test_{trg}$ to indicate the pre-translation of the test set (in the target-side language). The outputs of these two executions can be combined into one training set to build a model that produces better translations than models built using FDA having only $test_{src}$ as input.

Considering both the source side and target side of the parallel sentences as selection criteria is especially useful when using a corpus that includes sentences from subtitles in different languages. There are particular problems concerning parallel sentences comprising subtitles. For example, both sentences in the source and target side are limited to be displayed in the same time window (assuming they are synchronized). As the length of the same sentences in different languages can be different, this may causes the longest one to be rephrased, split in two, or have words omitted so it meets the time requirement.

In our work we use an approximated, synthetic target-side using a technique we call pre-translation. One way to
look at this is as a form of synthetic-data generation. As such it is somewhat reminiscent of synthetic source-data generation
using a target-to-source translation model, a technique known as back-translation introduced by Sennrich et al.(2016) \cite{sennrich2016improving}.

\section{Related Work}

%\subsection{Neural Machine Translation}

%We use neural machine translation \citep{kalchbrenner-blunsom:2013:EMNLP, cho-EtAl:2014:EMNLP2014} in the form of sequence-to-sequence models \citep{Sutskever:2014} based on recurrent neural networks \citep{DBLP:journals/corr/BahdanauCB14, luong-pham-manning:2015:EMNLP}.

%\subsection{Data Selection Techniques}
%\citep{poncelasextending,poncelas2017applying}

Data selection techniques aim to select a subset of data such that the models trained on that subset perform better. There are multiple approaches to achieve those improvements, such as domain adaptation or noise reduction approaches \citep{eetemadi2015survey}.

Methods based on domain adaptation include the work of Moore and Lewis (2010) \citet{moore2010intelligent}, who propose to use language models (LM) to select data. An LM is a distribution over sequences of words in a monolingual text, and is often used by SMT systems to model the fluency of the outputs. Given a string $s$ and a language model $LM_d$,  $H_d(s)$ is the entropy of the distribution of $s$ according to $LM_d$.
% In \textcolor{red}{the work by Moore and Lewis,} they build an in-domain language model $LM_I$ and an out-of-domain language model $LM_O$, so they can measure, for each sentence $s$, how likely is to be in-domain by computing the entropy difference $[H_{I}(s)-H_{O}(s)]$. 

Moore and Lewis build an in-domain language model $LM_I$ and an out-of-domain language model $LM_O$, and determine how likely each sentence $s$ is to be in-domain by computing the entropy difference $[H_{I}(s)-H_{O}(s)]$. 
Axelrod et al. (2011) \cite{axelrod2011domain} extend the method by using LMs in both the source-side and target-side languages, defining the bilingual cross-entropy difference.

Another method, proposed by van der Wees et al. (2017) \citet{van2017dynamic}, is to gradually remove out-of-domain sentences each $\eta$ epochs when training the NMT model.

In our work, we select data that is similar to $test_{src}$ (and so, more relevant for use as training data). 
%Previous research on selecting data considering the test set includes the work of \cite{lu2007improving} where they retrieve the sentences based on the TF-IDF \citep{tfidf1973} similarity measure. Another approach for applying in NMT is presented in the work of \citet{li2018one} were they fine-tune a pre-built NMT model using training data selected based on $test_{src}$. They use similarity measures, such as Levenshtein \citep{levenshtein1966binary} or the cosine similarity of the average of the word embeddings, \citep{mikolov2013distributed}. 
Previous research on selecting data considering the test set includes the work of Li et. al. (2018) \citet{li2018one} where they fine-tune a pre-built NMT model using training data selected based on $test_{src}$. They use similarity measures, such as Levenshtein distance \citep{levenshtein1966binary} or the cosine similarity of the average of the word embeddings, \citep{mikolov2013distributed}.

The method that we use to select data is FDA \citep{biccici2011instance,bicici2015parfda,biccici2015optimizing}, which has already proven to be useful in SMT \citep{biccici2013feature,poncelasextending,poncelas2017applying} and NMT \citep{poncelas2018feature}. Selecting a small subset of sentences from a parallel corpus using FDA is enough to train SMT systems that perform better than systems trained using the whole parallel corpus.

FDA takes as input a set of parallel sentences $U$ and a seed (generally the $test_{src}$). Given $U$ and the seed, FDA retrieves an ordered sequence of sentences $L$ from $U$. Sentences are ordered based on the amount of {\em n}-grams they share with the seed, with more shared {\em n}-grams meaning higher preference, while also considering the variability of the {\em n}-grams in the selected sentences.

% As output, it retrieves a sequence of sentences $L$ which contains sentences from $U$ but ordered so the top sentences are the ones that share the \textcolor{red}{most} {\em n}-grams with the seed, considering also the variability, selecting different {\em n}-grams.

The algorithm initializes $L$ as a void sequence and iteratively selects one sentence $s \in U-L$  and appends it to $L$. The sentence $s$ to select at each step is the one most relevant to $test_{src}$, based on the number of {\em n}-grams that $s$ shares with the $test_{src}$. The score of the relevance is computed as in \eqref{eq:fda_sentencescore}:

\begin{equation}\label{eq:fda_sentencescore}
score(s)=\frac{\sum_{f \in F_s} 0.5^{C_L(f)}}{\text{\# words in s}}
\end{equation}

\noindent where $F_s$ is the set of {\em n}-grams present in $s$ and $test_{src}$ (by default the order of the {\em n}-grams ranges from 1 to 3). $C_L(f)$ is the count of the {\em n}-gram $f$ in the sequence $L$ of selected sentences. Including $C_L(f)$ in the computation of the score causes the algorithm to penalize {\em n}-grams that have been selected several times, and hence favouring the selection of sentences that contain new {\em n}-grams.

%\marginpar{Sometimes you use arbitrary capitalization such as in ``parallel text''}

% \section{Using $test_{trg}$ in FDA}
\section{Using an Approximated Test Target-side}

\begin{figure*}[hbt]
\includegraphics[width=12cm, height=9cm]{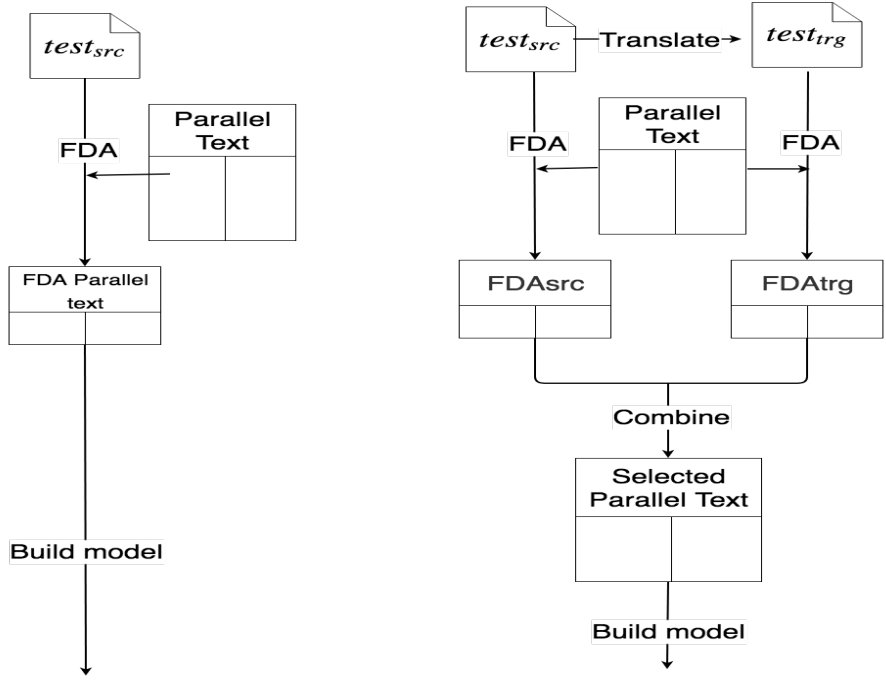}
\centering
\caption{Pipeline of the traditional usage of FDA (left) and pipeline of our proposal, using the target-side (right).
\label{src_trg_FDA_pipeline}}
\end{figure*}

FDA uses $test_{src}$ as seed to retrieve a subset from a set of parallel sentences. In order to retrieve the sentences it scores the {\em n}-grams of $test_{src}$ (source-side language). We show the pipeline of usage of FDA on the left side of Figure \ref{src_trg_FDA_pipeline}. Here, the files $test_{src}$ and \textit{parallel text} are used as input, and FDA retrieves a subset of the sentences to be used for building a model that is adapted to $test_{src}$.

% In our approach, we propose to use FDA using $test_{src}$ and $test_{trg}$ to obtain the set of sentences from the \textit{FDA Parallel text}.
% (using both source and target-side languages). 

We propose to use both the test source-side $test_{src}$ and the approximated test target-side $test_{trg}$ as features in FDA, when selecting the set of sentences from the  parallel text.

We show the pipeline of our approach on the right side of Figure \ref{src_trg_FDA_pipeline}. First, $test_{src}$ is translated (\textit{translate} step). Then, using FDA, we select a subset of parallel sentences given: (a) $test_{src}$ as seed ($FDA_{src}$), and (b) $test_{trg}$ as seed ($FDA_{trg}$). These two sets can be combined into one set which serves as training data to build an MT model.

In the following subsections we explain in more detail two issues that are yet unanswered in the pipeline : (1) how to build $test_{trg}$ (addressed in Section \ref{sec:pretranslate_testset}), and (2) how to combine the outputs of FDA (addressed in Section \ref{sec:combine_fda_output}). 

\subsection{Pre-Translation of $test_{src}$}
\label{sec:pretranslate_testset}

The first step in our approach consists of building $test_{trg}$ (\textit{translate} step on the right side of Figure \ref{src_trg_FDA_pipeline}) so it can be used as the seed to extract parallel sentences using the target side. In order to perform this pre-translation we need to build a model, which we refer to as the \textit{initial model}.

There are several approaches to build the \textit{initial model}, such as using SMT or NMT. These models can be trained using the full training data or subsets (such as randomly sampled, selected according to a particular domain, etc.). In this work we use an NMT model built with the full training data. 

%using a randomly sampled subset of training data. The amount of data we use to build this model is relevant for achieving a good performance. Using a small amount of data will lead to a small MT model that may produce a poor quality $test_{trg}$, and so the sentences retrieved in the following step will diverge from $test_{src}$.

%In previous experiments, we have discovered that model built in 1M sentences produce translations of quality that is acceptable. In Table \ref{table:baseline_results_1M} we show the translation output of the model built in 1M sentences.

\subsection{Combining FDA outputs}
\label{sec:combine_fda_output}

%The way of combining the outputs of $FDA_{src}$ and $FDA_{trg}$ we are exploring is the concatenation of them in order to build a data set of $N$ parallel sentences.

In order to combine the sentences of $FDA_{src}$ and $FDA_{trg}$ into one training set of N sentences, various strategies are possible such as retrieving the intersection or the union of sentences. In this work we explore the strategy of concatenating both outputs (allowing the repetition of sentences) and propose as future work alternative methods for merging both parallel datasets.

The outputs of $FDA_{src}$ and $FDA_{trg}$ can be seen as an ordered sequence of sentences as in equation \eqref{eq:FDA_src_seq} and equation \eqref{eq:FDA_trg_seq}:

\begin{equation}\label{eq:FDA_src_seq}
FDA_{src}=(s_1^{(src)},s_2^{(src)},s_3^{(src)},...s_{N}^{(src)})
\end{equation}

\begin{equation}\label{eq:FDA_trg_seq}
FDA_{trg}=(s_1^{(trg)},s_2^{(trg)},s_3^{(trg)},...s_{N}^{(trg)})
\end{equation}

In order to obtain a training set that combines the outputs of $FDA_{src}$ and $FDA_{trg}$, we concatenate the top sentences of each subset to obtain a new list of sentences of size $N$, as in equation \eqref{eq:FDA_comb_seq}
\begin{equation}\label{eq:FDA_comb_seq}
FDA=(s_1^{(src)},...s_{N*\alpha}^{(src)},s_1^{(trg)},...s_{N* (1-\alpha)}^{(trg)})
\end{equation}

%$\alpha \in [0,1]$
\noindent where $0 \leq \alpha \leq 1$ indicates the proportion of sentences that are selected from $FDA_{src}$ and $FDA_{trg}$.

Note that some of the sentences may be replicated; it may happen that $s_i^{(src)}=s_j^{(trg)}$, i.e. those that have been retrieved by both executions FDA. In this work we decided to keep the duplicates as it may be beneficial to oversample those sentences in which there is an agreement of both executions of FDA. However, we propose as future work to investigate the effect of removing those duplicate sentences.

%We assume that those sentences that are duplicated (those that have been retrieved by both executions FDA) indicate that they are relevant for $test_{src}$, so we assume it is beneficial to keep both of them. However, we propose as future work to investigate the effect of removing those duplicate sentences.

%Note that some of the sentences may be replicated, it may happen that $s_i^{(src)}=s_j^{(trg)}$. We assume that those sentences that are duplicated (those that have been retrieved by both executions FDA) indicate that they are relevant for $test_{src}$, so we assume it is beneficial to keep both of them. However, we propose as future work to investigate the effect of removing those duplicate sentences.

The core of our approach is combining the outputs of the two executions of FDA (using the test and translated sets). Given the concatenation method presented in this section, the outputs can be classified as one of the three options:

%We explore different possibilities of combining $FDA_{src}$ and $FDA_{trg}$:

\begin{itemize}
\item Source-side only: use only the output of $FDA_{src}$ for building the model. It is the configuration where $\alpha=1$ in Equation \eqref{eq:FDA_comb_seq}, which is equivalent to the traditional procedure of using FDA (left side of Figure \ref{src_trg_FDA_pipeline}, so we use this approach as the baseline.
\item Target-side only: use the output of $FDA_{trg}$ for building the model, which is the configuration where $\alpha=0$ in Equation \eqref{eq:FDA_comb_seq}.
\item Source-and-target-side: combine $FDA_{src}$ and $FDA_{trg}$. This is the configuration where different values of $\alpha$ in equation \eqref{eq:FDA_comb_seq} are set. In our work we explore the values $\alpha=0.25$, $\alpha=0.50$ and $\alpha=0.75$.
\end{itemize}

\section{Experiments}
\label{sec:experiments}

%The pipeline presented in this work can be seen from different perspectives. We can assume the initial model has been built by ourselves (so we have the training data of that model available) or consider the initial model to be an external resource. For this reason, we perform the experiment in these two different scenarios:

%\begin{enumerate}
%    \item \textit{Initial model as internal resource}: The problem is seen as a way of organizing the data in the pipeline to achieve the best results. In this context we allow FDA to select data that was used for building the initial model and we use the initial model as an intermediate step in the pipeline of a definitive model.
%    \item  \textit{Initial model as external resource}: We assume the initial model is an external resource, where we are not provided with the data used for building it and all we can do is to use it as black-box along with our data to build other models (FDA cannot select data that was used for building the initial model.)
%\end{enumerate}

\subsection{Experimental Settings}

We experiment with models for German-to-English direction. The parallel data used for the experiments is the training data provided in the {WMT 2015 \citep{bojar-EtAl:2015:WMT} (4.5M sentence pairs, 225M words). The dev set of the NMT models (both the initial model and those trained using the selected datasets) are 5K randomly sampled sentences from development sets from previous years. All the models presented here are evaluated using the same test set which comprises documents provided in WMT 2015 translation task as $test_{src}$.

In order to build the NMT models we use OpenNMT-py,
%\footnote{\url{https://github.com/OpenNMT/OpenNMT-py}}, 
which is the Pytorch port of OpenNMT \citep{opennmt}. All the NMT models we build use the same settings (we only change the training data used to build them). The value parameters are the default ones of OpenNMT-py (i.e. 2-layer LSTM with 500 hidden  units, vocabulary size of 50000 words for each language). All the models are executed for 13 Epochs.

In the experiments we build models with the data selected by using $FDA_{src}$ and $FDA_{trg}$. We explore selecting different sizes of selected data: 500K, 1M and 2M sentence pairs.

\section{Results}

\begin{table}[!htbp]
\centering
\begin{center}
%\begin{tabular}{ |p{2.5cm}|P{2.5cm}||P{2.5cm}|P{2.5cm}|P{2.5cm}|}
\begin{tabular}{ |p{2cm}|p{2cm}|}
\hline
& baseline \\
\hline									
BLEU	&	0.2474	\\
%NIST	&	6.9342	\\
TER 	&	0.5525	\\
METEOR	&	0.2798	\\
CHRF3	&	48.9473	\\
%CHRF1	&	50.4082	\\
\hline
\end{tabular}
\caption{ 
Results of the model trained with all available training data; also the no-FDA baseline.
\label{table:NMTFDA_allData}}
\end{center}
\end{table}

First, we show in Table \ref{table:NMTFDA_allData} the quality of the pre-translated $test_{trg}$. This has been produced by the \textit{initial model}, an NMT model trained with all training data. This result also serves as a no-FDA baseline to asses the benefit of using FDA in general with.
% The evaluation metrics presented in Table \ref{table:NMTFDA_allData} give an estimation of how similar is the output of the model compared to a human-translated reference. 

The evaluation metrics presented in Table \ref{table:NMTFDA_allData} give an estimation of the similarity between the model output and a human-translated reference.
The evaluation metrics we use are: BLEU \citep{papineni2002bleu}, 
%NIST \citep{doddington2002automatic}, 
TER \citep{snover2006study}, METEOR \citep{banerjee2005meteor} 
and CHRF \citep{popovic2015chrf}.

%The results of the final model are shown on Table \ref{table:results_srctrag_FDA_all} (\textit{Initial model as internal resource} scenario) and Table \ref{table:results_srctrag_FDA_new} (\textit{Initial model as external resource} scenario). These tables show the evaluation scores obtained by the models. 

The results of the models are shown in Table \ref{table:results_srctrag_FDA}. The columns show the different configurations used to build the set of selected sentences (i.e. the value of $\alpha$ in equation \eqref{eq:FDA_comb_seq} used). This means that the column $\alpha=0.75$ shows the results of the model trained with the sentences from the top-750K sentences of $FDA_{src}$ and the top-250K sentences of $FDA_{trg}$.

First, one may wonder whether FDA data selection is at all helpful? Comparing the scores in Table \ref{table:results_srctrag_FDA} to the baseline system trained on all data in Table \ref{table:NMTFDA_allData}, we see that all FDA systems outperform it, with the best one obtaining more than 1.5 BLEU points improvement (a relative improvement of 6\%).

We have marked in bold the scores that outperform the second baseline: FDA applied using $test_{src}$ only (i.e the configuration using $FDA_{src}$ and $\alpha=1$), as proposed in \citep{poncelas2018feature}, and computed the statistical significance (marked with an asterisk) with multeval \citep{clark2011better} for BLEU, TER and METEOR when compared to the baseline at level p=0.01 using bootstrap resampling \citep{koehn04}.

\begin{table*}[!htbp]
\centering
%\small
\begin{center}
\begin{tabular}{ |p{0.5cm}|p{1.5cm}|p{1.5cm}|p{1.5cm}|p{1.5cm}|p{1.5cm}|p{1.5cm}|}
\hline
 &&	$\alpha=1$  & $\alpha=0.75$ & $\alpha=0.50$ & $\alpha=0.25$ & $\alpha=0$ \\
\hline											
\multirow{4}{*}{\rotatebox[origin=c]{90}{\centering 500K lines}}
&BLEU	&	0.2517	&	\bf0.2542	&	\bf0.2543	&	\bf0.2534	&	0.2441	\\
%&NIST	&	7.0764	&	\bf7.1196	&	\bf7.0960	&	\bf7.1014	&	6.8898	\\
&TER	&	0.5601	&	\bf0.5521*	&	\bf0.5563	&	\bf0.5544	&	0.5628	\\
&METEOR	&	0.2886	&	\bf0.2895	&	0.2882	&	\bf0.2888	&	0.2789	\\
&CHRF3	&	49.8314	&	\bf50.0915	&	\bf49.8898	&	\bf49.9074	&	48.7796	\\
%&CHRF1	&	51.4945	&	\bf51.7894	&	\bf51.6422	&	\bf51.5453	&	50.1906	\\
\hline											
\multirow{4}{*}{\rotatebox[origin=c]{90}{\centering 1M lines}}		
&BLEU	&	0.256	&	\bf0.2627*	&	\bf0.2595	&	\bf0.2600*	&	0.2496	\\
%&NIST	&	7.098	&	\bf7.2215	&	\bf7.1823	&	\bf7.1931	&	7.0056	\\
&TER	&	0.5497	&	\bf0.5455*	&	\bf0.5462	&	\bf0.5493*	&	0.5534	\\
&METEOR	&	0.2886	&	\bf0.2920*	&	\bf0.2921*	&	\bf0.2918*	&	0.2833	\\
&CHRF3	&	50.0932	&	\bf50.6273	&	\bf50.5226	&	\bf50.5682	&	49.5192	\\
%&CHRF1	&	51.7129	&	\bf52.2635	&	\bf52.1180	&	\bf52.0730	&	50.8200	\\

\hline											
\multirow{4}{*}{\rotatebox[origin=c]{90}{\centering 2M lines}}											
&BLEU	&	0.2585	&	\bf0.2610	&	0.2580	&	\bf0.2614	&	0.2547	\\
%&NIST	&	7.1278	&	\bf7.2080	&	\bf7.1542	&	\bf7.2005	&	7.0666	\\
&TER	&	0.5454	&	\bf0.5429	&	0.5465	&	\bf0.5437	&	0.5496	\\
&METEOR	&	0.2894	&	\bf0.2923*	&	\bf0.2903	&	\bf0.2927*	&	0.2852	\\
&CHRF3	&	50.095	&	\bf50.5582	&	\bf50.2431	&	\bf50.5487	&	49.7838	\\
%&CHRF1	&	51.6795	&	\bf52.0516	&	\bf51.7644	&	\bf52.0237	&	51.1835	\\
\hline	
\end{tabular}
\caption{Results of the models using different sizes of $FDA_{src}$ and $FDA_{trg}$. }
\label{table:results_srctrag_FDA}
\end{center}
\end{table*}

%\subsection{\textit{Initial model as internal resource} and \textit{Initial model as external resource}}

%When comparing \textit{Initial model as internal resource} scenario (Table \ref{table:results_srctrag_FDA_all}) with  \textit{Initial model as external resource} (Table \ref{table:results_srctrag_FDA_new}) scenario we see that, not surprisingly, the best results are obtained in the first scenario because, as explained in Section \ref{sec:experiments}, there are more sentence candidates to be selected.

%using larger sets of sentences to extract data from obtain better results. We see that the evaluation scores obtained in Table \ref{table:results_srctrag_FDA_all} are better than those in Table \ref{table:results_srctrag_FDA_new}. 

%Both in Table \ref{table:results_srctrag_FDA_all} and Table \ref{table:results_srctrag_FDA_new} we find that those experiment in which $FDA_{src}$ and $FDA_{trg}$ are combined ($\alpha=0.75$, $\alpha=0.50$ and $\alpha=0.25$ columns) obtain better results than using $FDA_{src}$ alone ($\alpha=1$ column). However in Table \ref{table:results_srctrag_FDA_new} we find that adding sentences from $FDA_{trg}$ achieves statistical significant improvements for the models trained in 1M and 2M sentence pairs. This indicates that using $test_{trg}$ for selecting data is specially useful when an external MT model is used for generating it.

\subsection{Ratio of data obtained using source and target side}
%\subsection{Using $FDA_{src}$ and $FDA_{trg}$}

%When comparing the outputs of $FDA_{src}$ and $FDA_{trg}$ we find that there are 349192 sentences in common (35\% of the data) for these execution using all data. 

%If we compare the performance of models built in the output of FDA using only the information from the test set (column $\alpha=1$ in Tables \ref{table:results_srctrag_FDA_all} and \ref{table:results_srctrag_FDA_new}) with FDA using only $test_{trg}$ (column $\alpha=0$ in Tables \ref{table:results_srctrag_FDA_all} and \ref{table:results_srctrag_FDA_new}). 

%We see that they obtain similar results. 

Intuitively, models built using the data selected based on $test_{trg}$ might perform worse than using $test_{src}$ only. $test_{trg}$ may contain errors produced by the machine-generated text, so an algorithm that bases the decision on that text may not select the best sentences. Indeed, this can be seen in the column $\alpha=0$ of Table \ref{table:results_srctrag_FDA}, where most of the scores are worse than those in column  $\alpha=1$.
%used as a second baseline.

%Indeed, relying fully on  $test_{trg}$ is not a good idea, as can be seen in the columns $\alpha=0$ of Table \ref{table:results_srctrag_FDA}. Most of the scores are worse than those in column  $\alpha=1$ used as a baseline. 

%The only exception to this are the sub-tables \textit{1M} and \textit{2M} in Table \ref{table:results_srctrag_FDA_new}, but none of these scores are statistically significant improvements.

%\textcolor{red}{
%On the other hand, $test_{src}$ also has its limitations as a selection criterion, since it is not guaranteed that correct target-side translations can always be found by relying on only the source-side for sentence selection. 
%}

On the other hand, using only $test_{src}$ as a selection criterion also has its limitations. While it guarantees the selected source sentences to be similar to $test_{src}$, it does not provide any information about the target side of the selected sentences. Therefore, it may still select sentences with target-side translations that are wrong or not suitable given the domain of the test-set, thereby hurting the final translation accuracy.

Using training data containing parallel sentences that are not an accurate translation of each other is a problem that can be encountered when using parallel sentences obtained from subtitles. Often, translation of subtitles needs to be adapted to meet length requirements (due to the restriction of time it is displayed on screen). We present some examples of sentences that are not accurately translated in Table \ref{table:examples}.

%, since it is not guaranteed that correct target-side translations can always be found by relying on only the source-side for sentence selection. 

We find that selecting sentences based both on $test_{src}$ and on $test_{trg}$ works better than using one selection criterion alone. Thus, using an approximated target side, even if imperfect, can help. The best performance is obtained using configurations that combine outputs of $FDA_{src}$ and $FDA_{trg}$ ($\alpha=0.75$, $\alpha=0.50$ and $\alpha=0.25$ columns). 
% We cannot find a value of $\alpha$ that clearly make the results improve over the other values. 

%\textcolor{red}{
%The best results are obtained for $\alpha=0.25$ using 1 million sentences for selection. This setting improves 1.1 BLEU point over the baseline that uses only the source side for selection in FDA, with smaller but still statistically significant improvements for METEOR and TER.
%}
The best results are obtained for $\alpha=0.75$ using 1 million sentences for selection. This setting improves 1.53 BLEU points over the no-FDA baseline (model trained with all data) and 0.67 BLEU points over the baseline that uses only the source side for selection in FDA.

%\textcolor{red}{
%When selecting 2 million sentences,  $\alpha=0.75$ performs relatively better than  $\alpha=0.25$, but overall the results are worse than when only 1 million sentences are selected.}
%}

\begin{table*}[!htbp]
\centering
%\small
\begin{center}
\begin{tabular}{ |p{7cm}|p{7cm}|p{1cm}|p{1cm}|}
\hline
German & English & \footnotesize pos $FDA_{src}$ & \footnotesize pos $FDA_{trg}$\\
\hline
nun gibt es kein Zurück mehr .	&	there is no going back now .	&	12	&	-\\
\hline
diese Zahl ist mehr als doppelt so viel , als vor 10 Jahren .	&	famous pieces from the 19th century include those by Delacroix , Gauguin , Monet , Renoir and Corot .	&	50	&	-\\
\hline
diese Aufzählung ließe sich beliebig fortführen .	&	and I could continue .	&	-	&	63\\
\hline
bitte beachten Sie , dass Sie sich registrieren lassen müssen , um einen Zugang zu den detaillierten Außenhandelsdaten zu erhalten .	&	all data can be downloaded free of charge .	&	-	&	92\\
%\hline
%er ist verheiratet und hat zwei Kinder .	&	since then , he has had a long career on stage , in film and on television . he has also established himself as a singer and an author in recent years .	&	22	&	99434\\
%\hline
%he was born in Richmond , Virginia .	&	am 13. Januar 1990 wurde er Gouverneur von Virginia .	&	209752	&	2\\
%\hline
%do not hesitate to contact us .	&	für weitere Informationen stehen wir gerne zur Verfügung .	&	347931	&	14\\
\hline
\end{tabular}
\caption{Examples of sentences retrieved by $FDA_{src}$ and $FDA_{trg}$}
\label{table:examples}
\end{center}
\end{table*}

%trg
%er hat darüber nichts gewußt .	he did not know anything about it .
%doch das geschah nicht .	but it did not happen .
%mehr kann ich gegenwärtig nicht dazu sagen .	that is all I can say for the time being .
%dieses Phänomen wird sich also noch weiter verbreiten .	all of this means that the disease will spread .

In Table \ref{table:examples} we show examples of sentences that are exclusive outputs of $FDA_{src}$ or $FDA_{trg}$. These examples give an indication about how including the output of $FDA_{trg}$ can benefit (or hurt) the quality of the selected data.

In the first row we see that the sentence \say{nun gibt es kein Zurück mehr .} has been selected by $FDA_{src}$ as it matches \say{kein Zurück mehr} in the input. According to this sentence, this {\em n}-gram should be translated as \say{no going back}. The translation found for \say{kein Zurück mehr} in $test_{trg}$ is \say{point where there is no return} (which, in addition, is closer to the reference \say{point of no return}) and hence $FDA_{trg}$ will use {\em n}-grams such as \say{point} or \say{no return} to retrieve sentences.

In the second row, we find an example of a sentence retrieved by $FDA_{src}$ whose translation is not accurate (this is easily noticeable as the names \say{Delacroix, Gauguin, Monet, Renoir and Corot} are not present in the English-side sentence). Including this sentence in the training data causes the quality to decrease and the models to perform worse. This problem is not exclusive of $FDA_{src}$, as in rows 3 and 4 we see the same problem happening in the output of $FDA_{trg}$.

Combining the outputs of $FDA_{src}$ and $FDA_{trg}$ causes the training data to be reinforced with sentences with relevant translations. Note that mixing the outputs of the two executions of FDA cause some sentence pairs to be replicated, as there is an overlap of the outputs. 

%In Figure \ref{FDAsrc_vs_FDAtrg} we see how the overlap increases the more sentences of $FDA_{trg}$ (horizontal axis) are selected. We see that in this particular case the increments are linear.

%the number of sentences that overlap with the top-500K, top-1M and top-2M outputs $FDA_{src}$ increases when selecting the top-N sentences of $FDA_{trg}$.

%\begin{figure*}[hbt]
%\includegraphics[width=15cm, height=8cm]{images_src_trg_FDA/FDAsrc_vs_FDAtrg.png}
%\centering
%\caption{Number of sentences that overlap with $FDA_{src}$ (top-500K, top-1M and top-2M outputs) when selecting different number of sentences of  $FDA_{trg}$.
%\label{FDAsrc_vs_FDAtrg}}
%\end{figure*}

%The amount of duplicated sentences in the training data of the models built (those presented in Table \ref{table:results_srctrag_FDA}) can be seen in Table \ref{table:FDA_src_trg_overlap}.

In Table \ref{table:FDA_src_trg_uniq} we indicate the amount of unique lines contained in the training data of the models (those presented in Table \ref{table:results_srctrag_FDA}). In the table we observe that the number of unique lines is high in all training sets. The proportion of unique lines ranges from 82\% to 94\%, which shows how $FDA_{src}$ and $FDA_{trg}$ retrieve different sentences mostly.

%\begin{table}[!htbp]
%\centering
%%\small
%\begin{center}
%\begin{tabular}{ |p{0.7cm}|p{2cm}|p{2cm}|p{2.2cm}|}
%\hline
% & $\alpha=0.75$ & $\alpha=0.50$ & $\alpha=0.25$  \\
%\hline
%500K 	&	96183 (19\%)	&	211002 (42\%)	&	346760 (69\%)	\\
%1M 	&	167105 (17\%)	&	386695 (39\%)	&	668520 (67\%)	\\
%2M 	&	245167 (12\%)	&	648754 (32\%)	&	1249036 (62\%)	\\
%\hline							
%\end{tabular}
%\caption{Number of sentences that overlap in $FDA_{src}$ and $FDA_{trg}$. }
%\label{table:FDA_src_trg_overlap}
%\end{center}
%\end{table}

\begin{table}[!htbp]
\centering
%\small
\begin{center}
\begin{tabular}{ |p{0.4cm}|p{2.1cm}|p{2.1cm}|p{2.1cm}|}
\hline
 & $\alpha=0.75$ & $\alpha=0.50$ & $\alpha=0.25$  \\
\hline
\footnotesize 500K 	&	 471753 (94\%)	&	460993 (92\%)	&	 471174 (94\%)	\\
\footnotesize 1M 	&	 918506 (92\%)	&	 886685 (89\%)	&	 917087 (92\%)	\\
\footnotesize 2M 	&	 1749015(87\%)	&	 1648727(82\%)	&	 1745142(87\%)	\\
\hline							
\end{tabular}
\caption{ Number of unique sentences in the training data. }
\label{table:FDA_src_trg_uniq}
\end{center}
\end{table}

%\textcolor{red}{
%We see that the larger the selected data is the more overlap there is between $FDA_{src}$ and $FDA_{trg}$ (the rows 2M are those with the highest overlap in each subtable of Table \ref{table:FDA_src_trg_overlap}). The highest observed overlap is the 2M row for $\alpha=0.25$. This is the configuration that produced the output measured in column $\alpha=0.25$ of subtable 2M in Table \ref{table:results_srctrag_FDA}. 
%}

%\textcolor{red}{
%We can see that this model outperforms those trained on data without duplicated instances ($\alpha=1$ and $\alpha=0$ columns of subtable 2M). This reveals that selection algorithms can be improved if certain sentence pairs are selected twice instead of adding new instances.
%}

When performing a column-wise comparison in Table \ref{table:FDA_src_trg_uniq}, we see how the number of unique lines is larger when the output of one of the FDA models dominates the training data ($\alpha=0.25$ or $\alpha=0.75$ columns) compared to those sets that contain the same amount of sentences extracted from $FDA_{src}$ and $FDA_{trg}$ (column $\alpha=0.50$).

We also see that the larger the amount of selected data, the more overlap exists between the two outputs (the proportional amount of unique lines is smaller). For example, in column $\alpha=0.50$, when 500K lines are selected, there are 92\% non-repeated lines, and this decreases to 82\% when selecting 2M lines. The same can be observed in the other columns. This indicates how the selected data using $FDA_{src}$ and $FDA_{trg}$ tend to be more similar the more sentences are retrieved.

\section{Conclusion and Future Work}

In this work, we explored a different pipeline in which FDA can be used. We discovered that using $test_{trg}$ (which is machine-generated) as the seed of FDA can improve the performance. 

In our experiments, we built models using training sets containing replicated instances of sentence pairs (as the output of the two runs of FDA, on the source-side and target-side, may overlap). This opens the door to exploring data selection algorithms allowing the repetition of selected instances.

In the future, we want to consider other procedures for combining the outputs of FDA, as we believe that other merging strategies may achieve better results. For example, considering both {\em n}-grams on the source and target side in combination (rather than two separate executions of FDA) may achieve better performance.

In addition, we want to explore the performance when using a different \textit{initial model}. Changing the initial model to produce the $test_{trg}$ causes $FDA_{trg}$ to have a different performance. We believe that using another dataset to build the initial NMT model (or even using different paradigms such as SMT or rule-based  MT) or choosing an initial model that is also closer to $test_{src}$ (e.g. using FDA to build the initial model) should boost the performance. Moreover, the use of several initial models allow us to perform concatenations of several outputs of FDA using different seeds.

Finally, we want to explore how data selection algorithms may improve when allowing the algorithm to select the same sentence pairs several times.

%changing the initial NMT model (or even using different paradigms such as SMT or rule-based  MT). We believe that using an initial model that it is also closer to $test_{src}$ (e.g. using FDA to build the initial model) should boost the performance. In addition, we want to explore how data selection algorithms may improve when allowing the algorithm to select several times the same sentence pairs.

\section{Acknowledgements}

This research has been supported by the ADAPT Centre for Digital Content Technology which is funded under the SFI Research Centres Programme (Grant 13/RC/2106) and is co-funded under the European Regional Development Fund.

 \noindent 
\includegraphics[width=1cm]{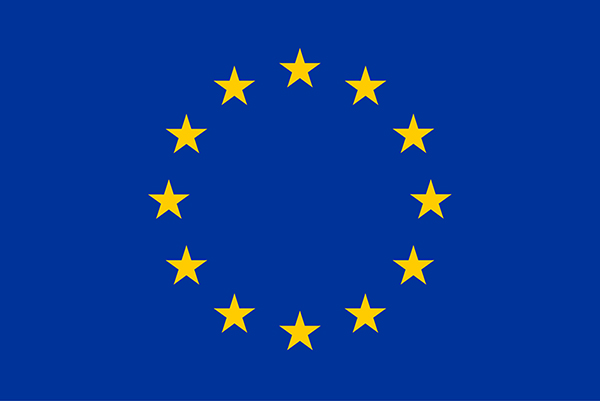}
This work has also received funding from the European Union’s Horizon 2020 research and innovation programme under the Marie Skłodowska-Curie grant agreement No 713567.

%\section{References}
%\FloatBarrier
%\clearpage
%\bibliographystyle{plainnat}
\bibliographystyle{IEEEtran}
\bibliography{bibl-no-url}
%\bibliography{bibl}
\end{document}